\documentclass[10pt,twocolumn,letterpaper]{article}

\usepackage{cvpr}              

\usepackage{graphicx}
\usepackage{booktabs}
\usepackage{cite}
\usepackage{amsmath,amssymb,amsfonts}
\usepackage{algorithmic}
\usepackage{textcomp}
\usepackage{xcolor}
\usepackage{dsfont}
\usepackage[accsupp]{axessibility}
\usepackage[pagebackref,breaklinks,colorlinks]{hyperref}

\usepackage[capitalize]{cleveref}
\crefname{section}{Sec.}{Secs.}
\Crefname{section}{Section}{Sections}
\Crefname{table}{Table}{Tables}
\crefname{table}{Tab.}{Tabs.}

\def\cvprPaperID{6322} 
\def\confName{CVPR}
\def\confYear{2022}

\makeatletter
\def\thanks#1{\protected@xdef\@thanks{\@thanks
        \protect\footnotetext{#1}}}
\makeatother

\begin{document}

\title{Cerberus Transformer: Joint Semantic, Affordance and Attribute Parsing
}

\author{Xiaoxue Chen$^{1}$, Tianyu Liu$^2$, Hao Zhao$^{3,4}$, Guyue Zhou$^1$, Ya-Qin Zhang$^1$ \\
	$^1$AIR, Tsinghua University $^2$HKUST $^3$Peking University $^4$Intel Labs\\
	{\tt\small \{chenxiaoxue, zhouguyue, zhangyaqin\}@air.tsinghua.edu.cn}\\
	{\tt\small tianyu.liu@connect.ust.hk}, {\tt\small zhao-hao@pku.edu.cn}, {\tt\small hao.zhao@intel.com}}
	

\maketitle

\begin{abstract}
Multi-task indoor scene understanding is widely considered as an intriguing formulation, as the affinity of different tasks may lead to improved performance. In this paper, we tackle the new problem of joint semantic, affordance and attribute parsing. However, successfully resolving it requires a model to capture long-range dependency, learn from weakly aligned data and properly balance sub-tasks during training. To this end, we propose an attention-based architecture named Cerberus and a tailored training framework. Our method effectively addresses aforementioned challenges and achieves state-of-the-art performance on all three tasks. Moreover, an in-depth analysis shows concept affinity consistent with human cognition, which inspires us to explore the possibility of weakly supervised learning. Surprisingly, Cerberus achieves strong results using only $0.1\%-1\%$ annotation. Visualizations further confirm that this success is credited to common attention maps across tasks. Code and models can be accessed at \url{https://github.com/OPEN-AIR-SUN/Cerberus}.
\end{abstract}

\section{Introduction}
\label{sec:intro}

\begin{figure}[t]
  \centering
  \includegraphics[width=1\linewidth]{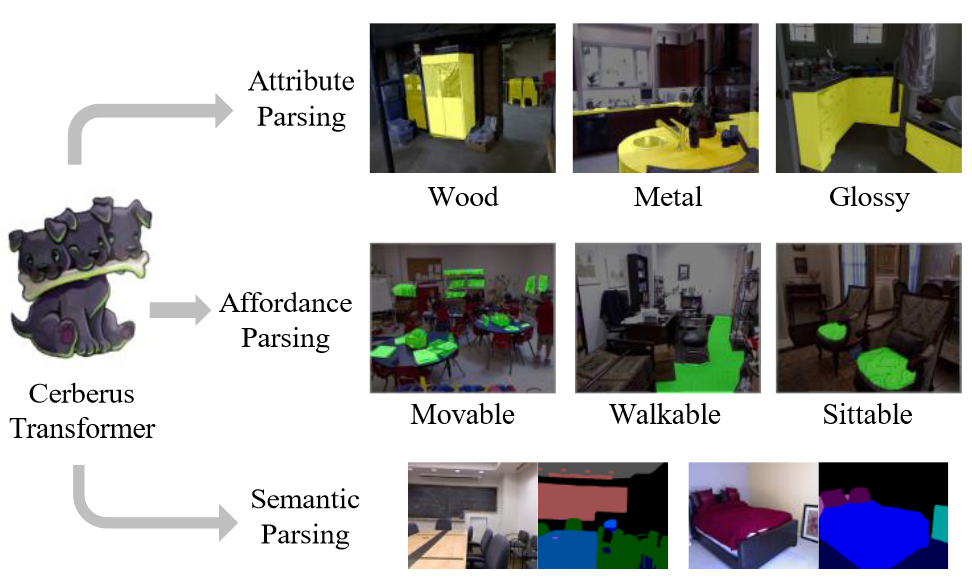}
  \caption{\textbf{Cerberus Transformer}. Given a single image, Cerberus parses attribute, affordance and semantics simultaneously. The cartoon is credited to https://www.redbubble.com/i/sticker/Baby-Cerberus-by-ArtOfBianca/48150266.EJUG5.}
  \label{fig:teaser} 
\end{figure}

Understanding indoor scenes is a fundamental computer vision topic, with many applications in intelligent robots and metaverse. To achieve a holistic understanding, many sub-tasks need to be addressed and it is widely believed and evidenced that jointly addressing them lead to more accurate results \cite{zhao2013scene}\cite{eigen2015predicting}\cite{song2017semantic}\cite{zamir2018taskonomy}\cite{han2020occuseg}. Different from former arts, we study a new and challenging formulation: joint semantic, affordance, and attribute parsing from a single image. As shown in Fig.~\ref{fig:teaser}, these three tasks cover a wide spectrum of human recognition and cognition abilities. The attribute of an object (like \emph{wood} or \emph{Glossy}) is a low-level physical property. The semantic category of a region (like \emph{floor} or \emph{sofa}) is a recognition-level concept. Affordance prediction (like \emph{movable} or \emph{walkable}) is a cognition-level problem. These three tasks are closely associated, since objects with specific semantics tend to have specific attribute or affordance. Parsing them jointly is a natural yet unexplored formulation. 

This new formulation brings both challenges and opportunities. In order to resolve three tasks with a single model, we need to learn shared representations that effectively serve all of them. Meanwhile, the representations are expected to model long range dependency in inputs in a principled manner. In order to simultaneously meet these two requirements, we resort to the transformer architecture \cite{vaswani2017attention}, which has a global receptive field at each layer. The proposed architecture is named as Cerberus.

Our formulation is challenged by another uncommon issue: weakly aligned data. During the historical development of scene understanding techniques, attribute \cite{zheng2014dense} and affordance \cite{roy2016multi} annotations are gradually added to the original NYUd2 semantic parsing dataset \cite{silberman2012indoor}. Unfortunately, their image-annotation pairs are only weakly aligned in the spatial domain. This is in contrast to former multi-task scene understanding methods which exploit aligned \emph{one-input-multi-output} datasets. To this end, we develop a tailored training framework that treats three datasets as separate sources and leverages a gradient projection technique on pre-computed task-wise gradient tensors. It unleashes the power of multi-task learning and boosts the quantitative results of all three tasks to a state-of-the-art level.

As mentioned before, opportunities come along with challenges. We first conduct in-depth analyses to investigate concept affinity in our three tasks. Interestingly, we observe concept affinity matrices well aligned with human cognitive commonsense. For example, if a pixel is predicted as \emph{floor}, naturally it should be labelled as \emph{walkable}. This finding inspires us to leverage task affinity for weakly supervised learning. During the training of Cerberus, we reduce the annotation amount of a specific sub-task to only $0.1\%-1\%$ and rely upon representations learnt by other sub-tasks. It is shown that Ceberus consistently out-performs baselines by significant margins in these settings. What's more, we visualize attention maps and validate that the capability of weakly-supervised learning is indeed enabled by shared attention weights. We argue this is a human-like learning feature: If one (e.g., an infant) knows what is floor, then she can learns where is walkable using very few examples.
  
We have following contributions: (1) We propose a novel multi-task dense prediction transformer named Cerberus, for joint semantic, affordance and attribute parsing in indoor scenes; (2) Cerberus achieves state-of-the-art results for all three tasks while requiring a single forward pass, with the help of a task weight balancing framework that learns from weakly-aligned data. (3) Via extensive analyses, we show that Cerberus learns task affinity consistent with human cognition and achieves strong weakly-supervised learning performance using only 0.1\% annotation.


\section{Related Works}

\textbf{Transformer} \cite{vaswani2017attention} has transformed natural language processing since its advent. Due to its strong power to model long-range dependency and capture contextual information, transformer has been proven effective for both 2D \cite{liu2021swin}\cite{ranftl2021vision} and 3D \cite{liu2021group}\cite{chen2022pq} scene understanding problems. Apart from this established advantage, we think transformer is well suited for another potential scenario: multi-task dense prediction. The intuition is that related tasks naturally share attention weights, e.g., \emph{floor} and \emph{walkable}. Interestingly, we validate this point using both strong weakly-supervised learning results and intuitive visualizations.

\textbf{Scene understanding} has long been addresed in a multi-task setting, even before the advent of deep learning. A joint probabilistic formulation can incorporate priors and allow physically more plausible understanding \cite{hedau2010thinking} \cite{schwing2013box}\cite{choi2015indoor}. Incorporating deep representations leads to compelling holistic understanding capabilities including layout, object and human \cite{chen2019holistic++}\cite{nie2020total3dunderstanding}\cite{zhang2021holistic}. Semantic scene completion naturally entangles reconstruction and semantic labelling \cite{song2017semantic}\cite{zhang2018efficient}\cite{zhang2019cascaded}\cite{chen20203d}. \cite{zhao2017physics} exploits semantics-layout concept affinity for effective representation learning from unbalanced data. \cite{zamir2020robust}\cite{mao2020multitask} demonstrate improved robustness of deep models via exploiting multi-task consistency. While semantics, affordance, and attribute serve as three fundamental tasks in scene understanding, previous works\cite{parikh2011relative}\cite{kumar2009attribute}\cite{patterson2012sun}\cite{do2018affordancenet}\cite{sawatzky2017weakly}\cite{shotton2009textonboost}\cite{ladicky2009associative} \cite{krahenbuhl2011efficient}\cite{sawatzky2019object} address them separately. To our knowledge, Cerberus for the first time addresses joint semantic, affordance, and attribute parsing in this large literature. New challenges addressed and new opportunities captured, as mentioned above, distinguish this study from former ones.

\section{Method}

In this paper, we aim at parsing semantics, affordance and attribute jointly. Semantics (e.g., \emph{sofa} or \emph{cabinet}) describes object/stuff categories in an indoor scene. Affordance means an object's capability to support a certain human action, for instance \emph{walkable} or \emph{sittable}. And attribute refers to object material like \emph{metal} or surface properties like \emph{shiny}. Via predicting these labels, an agent understands an indoor scene in a comprehensive manner. We define $\mathcal{O}= \left\{o_1,o_2,...,o_x\right\}$ as the semantic label set, $\mathcal{F} =\left\{f_1,f_2,...,f_{y}\right\}$ as the affordance label set, and $\mathcal{T} = \left\{t_1,t_2,...,t_z\right\}$ as the attribute label set. Given an image I, for each pixel $\rm I_i$, the task is formally stated as a mapping
\begin{align}
\rm I_i \rightarrow \mathcal{O} \times \mathcal{P}(\mathcal{F}) \times \mathcal{P}(\mathcal{T})
\end{align}
where $\mathcal{P}$ is the power-set operator, and $\times$ is the Cartesian product operator. This means each pixel corresponds to one semantic label, j affordance labels and k attribute labels, where $0\leq j\leq y, 0\leq k\leq z$.


\begin{figure*}[t]
  \centering
  \includegraphics[width=0.8\linewidth]{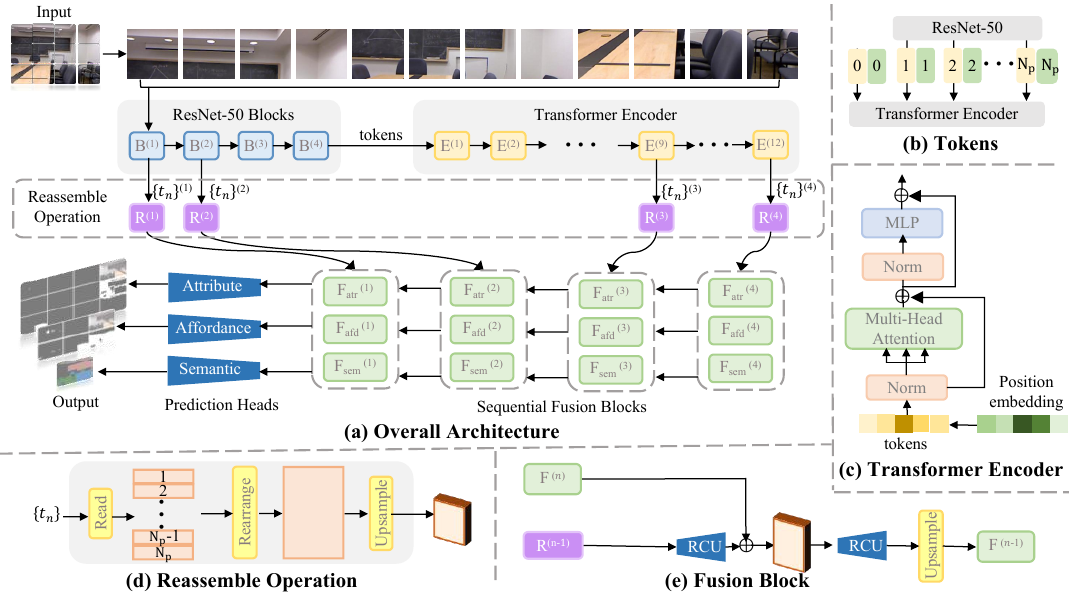}
  \caption{\textbf{Overall network architecture of Cerberus.} Given an image, ResNet-50 extract features from the input image to form a set of tokens. The tokens are processed by a transformer encoder and decoded by reassemble operations and fusion blocks. Through three prediction heads, the feature maps are turned into final attribute, affordance and semantic parsing results.}
  \label{fig:main} 
\end{figure*}

\subsection{Network Architecture of Cerberus}
Intuitively, these three tasks are not independent, e.g., pillows are intrinsically movable. We believe parsing them with a single network can improve performance by exploiting inductive biases between different tasks. However, what is the best architecture for multi-task dense prediction remains an open problem. Generic principles do exist: such an architecture should capture long-range dependency within visual inputs and learn shared representations that effectively serve several tasks. Our observation is that transformer well meets these two requirements: the attention operator has a global receptive field and learning attention focused on a region naturally facilitates representation sharing if different sub-task labels coexist in this region. Hence, we propose the first multi-task dense prediction transformer for joint semantic, affodance and attribute parsing, which is named Cerberus and depicted in Fig.\ref{fig:main}. 

\textbf{Transformer encoder}. Given an image of $H \times W$ pixels, we divide it into $N_p = \frac{HW}{p^2}$ non-overlapping square patches of size $p^2$. As illustrated in Fig.\ref{fig:main} (b), the set of patches is flattened into a vector of length $N_p$ then passed through a ResNet-50 backbone to form $N_p$ embeddings. The embeddings are denoted as a set of tokens: $\left\{t_n\right\}, n=1,...,N_p$. Learnable position embeddings are concatenated with the tokens to retain positional information. Following \cite{ranftl2021vision}, an extra learnable token $t_0$ is added to the sequence, which serves for attention visualization in Fig.~\ref{fig:attention}. It aggregates information from the entire sequence and is named as a readout token. All the  $N_p+1$ tokens are then fed into sequential blocks of multi-head self-attention, which learn shared representations for different tasks.

\textbf{Reassemble operation}. After processing a set of tokens $\left\{t_n\right\}, n=0,...,N_p$  with transformer encoder, we then assemble them into image-like feature representations at various resolutions which is illustrated in Fig.\ref{fig:main} (d). 

\begin{figure*}[t]
	\centering
	\includegraphics[width=0.75\linewidth]{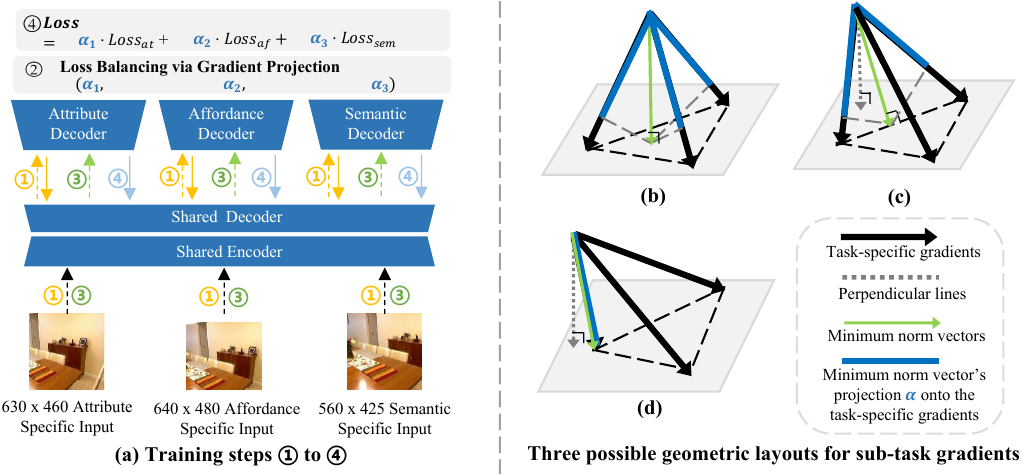}
	\caption{The illustration of \textbf{training framework} (left) and layouts of \textbf{gradient vectors} (right). }
	\label{fig:moo} 
\end{figure*}

First, We get $N_p$ embeddings by concatenating $t_0$ to all other tokens and project the embeddings to D-dimensional features using a fully connected layer. Then, we rearrange the new $ N_p$ features by placing them according to the position of the initial patch and get a feature map $\rm F_{rearrange} \in \mathbb{R}^{\frac{H}{p} \times \frac{W}{p} \times D}$. Next, we use a spatial unsampling layer to resize $\rm F_{rearrange}$ to $\rm F_{upsample} \in \mathbb{R}^{\frac{H}{s} \times \frac{W}{s} \times \hat{D}}$. We reassemble tokens from the outputs of four different stages (the first and second ResNet-50 blocks, layer 9 and layer 12 of transformer encoder) into four image-like representations with different resolutions.

\textbf{Fusion block}. After generating four feature maps from aforementioned stages, Cerberus uses RefineNet-style\cite{lin2017refinenet} feature fusion blocks to progressively upsample them. The fusion block is depicted in Fig.\ref{fig:main} (e). In the $n^{th}$ fusion stage, We use a residual convolutional unit (RCU) to process the reassembled feature $\rm R_{n-1}$ first, then fuse it with the previous feature $\rm F_{n}$ via another RCU after element-wise summation. Then we upsample the result by a factor of two and get the new fused feature map $\rm F_{n-1}$. We use the final fused feature map to generate task-specific predictions.

\textbf{Prediction head}. 
We use three separate prediction heads to produce the final dense prediction results. Each head is composed of two parts: (1) a fully connected layer to generate the semantic, affordance or attribute map, (2) an interpolation function to upsample the predicted map to the original image resolution. For affordance and attribute, we get maps of size $ \rm y \times H \times W$ and  $ \rm z \times H \times W$ respectively, where y and z are the number of label classes. And for semantics, the size of predicted map is $\rm  H \times W$, where each pixel corresponds to a semantic class. We use y binary cross entropy losses to supervise affordance, z binary cross entropy losses for attribute and a x-way cross entropy loss for semantics.

\subsection{Weakly-aligned Training with Optimal Weights}

\textbf{Motivation} How to train our multi-task dense prediction transformer in an effective manner? A straightforward idea is to use a naïve combination of different task losses:
\begin{align}
\mathcal{L}_{\rm multi\_task} = \sum_{t=1}^{T}w_t \mathcal{L}_{\rm t} \left(\theta\right)
\end{align}
T is the number of tasks, $w_t$ is the loss weights of tasks t and $\theta$ is the parameters of the network. As shown in former studies (e.g., Fig.2 in \cite{kendall2018multi}), the performance of the model is sensitive to the selection of weights. Tuning these weights manually is difficult and expensive. Moreover, the optimal weights might change during the training process, which is verified in our experiment (Fig.~\ref{fig:Weights}).

We are faced with another challenge: weakly aligned data. Though we use the same dataset to train semantics, affordance and attribute, the annotation of the three tasks have a spatial shift problem. For example, in Fig.\ref{fig:moo} (a), the input images for three tasks are taken from the same scene, however they are not strictly aligned and even have different resolutions. This means we can't use the one-input-multi-output training paradigms like \cite{sener2018multi}, but have to do forward propagation for each task. To resolve this issue while avoiding manual weight tuning, we resort to the original MGDA formulation \cite{desideri2012multiple} which is naturally compatible.

\textbf{Preliminaries} A solution $\theta_1$ dominates another solution $\rm \theta_2$ if $\rm \forall t$, $\mathcal{L}_{\rm t} \left(\theta_1\right) \leq \mathcal{L}_{\rm t} \left(\theta_2\right)$  and ${\rm \exists t}$, $ \mathcal{L}_{\rm t} \left(\theta_1\right) \textless \mathcal{L}_{\rm t} \left(\theta_2\right)$. We call a solution $\theta^{*}$ Pareto-optimal, if there is no other solution dominates  $\theta^{*}$. And a solution $\theta^{*}$ is said to be Pareto-stationary, if:
\begin{align}
\sum_{t=1}^{T} \alpha_t \nabla \mathcal{L}_{\rm t} \left(\theta^{*}\right) = 0,   
st.\rm \quad \sum_{t=1}^{T} \alpha_t = 1,\alpha_t \geq 0,\forall t. 
\end{align}
$ \nabla \mathcal{L}_{\rm t} \left(\theta^{*}\right)$ is the gradient of $\mathcal{L}_{\rm t} \left(\theta^{*}\right)$. If solution $\theta^{*}$ is Pareto-optimal, it is Pareto-stationary, but the reverse isn't always true\cite{desideri2012multiple}.  We consider the following optimization problem for Pareto-stationary:
\begin{align}
min_{\alpha_1...\alpha_T} \|\sum_{t=1}^{T} \alpha_t \nabla \mathcal{L}_{\rm t} \left(\theta \right)\| _2 = 0, \label{moo} \\
st. \rm \quad \sum_{t=1}^{T} \alpha_t = 1,\alpha_t \geq 0,\forall t. 
\end{align}
This is equivalent to finding the vector of minimum norm in the convex hull of the input gradient set. 

As for our attention-based model, the effectiveness of this optimization scheme is not clear, since self-attention instead of network parameters play the major role in a transformer. We take the solution of Eq.\ref{moo} as \textbf{optimal weights} to train our multi-task transformer, and empirical evidence shows that they effectively balance Cerberus (Tab.~\ref{tab:moo}). 

\textbf{Formulation} Considering the case of Cerberus, T is equal to 3. For notational simplicity, we denote $\nabla \mathcal{L}_{1} \left(\theta \right)$ as $g_1$, $\nabla \mathcal{L}_{2} \left(\theta \right)$ as $g_2$ and $\nabla \mathcal{L}_{3} \left(\theta \right)$ as $g_3$. The geometric illustration is shown in Fig. \ref{fig:moo} (right). The  minimum norm vector $w$ is either perpendicular to the convex hull or in an boundary case. If the minimum norm is a perpendicular vector which is illustrated in Fig. \ref{fig:moo} (b), we have:
\begin{align}
w = \alpha_1 g_1 + \alpha_2 g_2 + \alpha_3 g_3,\\
w \perp \left( g_1  - g_2 \right), w \perp \left( g_1  - g_3 \right).
\end{align}
which is equal to solving three ternary linear equations with constraints that interested variables are greater than zero:
\begin{align}
\begin{bmatrix} 
 g_1^{\rm T}  - g_2^{\rm T} \\
 g_1^{\rm T}  - g_3^{\rm T} \\
\end{bmatrix} \begin{bmatrix}
g_1 & g_2 & g_3\\
\end{bmatrix} 
\begin{bmatrix}
\alpha_1 \\ \alpha_2 \\ \alpha_3\\
\end{bmatrix} = 0,\\
st. \rm \quad \sum_{t=1}^{3} \alpha_t = 1,\alpha_t \geq 0,\forall t. 
\end{align}
If there is an analytical solution $\alpha_1^*,\alpha_2^*,\alpha_3^*$, the minimum norm vector is $w = \alpha_1^* g_1 + \alpha_2^* g_2 + \alpha_3^* g_3$. Otherwise, the minimum norm vector points to an edge, and it must be a convex combination of two gradient-vectors, which is illustrated in Fig.\ref{fig:moo} (c). Choose two gradient vectors with smaller norms, for instance $g_1$ and $g_2$, then there is:
\begin{align}
w = \alpha_1 g_1 + \alpha_2 g_2, \quad w \perp \left( g_1  - g_2 \right).
\end{align}
and the analytical solution is:
\begin{align}
\alpha_1^* = \frac{\left(g_2-g_1\right)^{\rm T}g_2}{\|g_2-g_1\|^2}
\end{align}
if $0\textless \alpha_1^*\textless 1$, then $w = \alpha_1^* g_1 + (1-\alpha_1^*) g_2$, otherwise the minimum norm vector points to a vertex which is depicted in Fig.\ref{fig:moo} (d). If $\alpha_1^*\geq1$, $w = g_1$, else if $\alpha_1^*\leq0$, $w = g_2$.

After solving the gradient-based problem Eq.(\ref{moo}), we get the optimal weights $\alpha_1^*,\alpha_2^*,\alpha_3^*$  for Cerberus:
\begin{align}
\mathcal{L}_{\rm Cerberus} = \alpha_1^* \mathcal{L}_{\rm at}+\alpha_2^* \mathcal{L}_{\rm af} + \alpha_3^* \mathcal{L}_{\rm sem} 
\end{align}
 $\mathcal{L}_{\rm at}$ is the loss for attribute, $\mathcal{L}_{\rm af}$ is the loss for affordance and $\mathcal{L}_{\rm sem}$ is the loss for semantics. With this joint loss, our model gradually converges to a Pareto-stationary solution. 
 
\textbf{Implementation}. To resolve weakly-aligned data and collect gradients for Eq.\ref{moo}, we propose a tailored training framework which  is demonstrated in Fig.\ref{fig:moo} (a). There is a forward propagation for each task, which is followed by a backward pass to calculate the task-specific gradient (step 1). Then we solve the gradient-based problem and update the optimal weights (step 2). Afterwards, we do forward passes again (step 3) to get prediction results. Finally, we calculate the joint loss with optimal weights and update the network with backward propagation (step 4). 

\begin{table}
	\centering
	\begin{tabular}{@{}lc@{}}
		\toprule
		\textbf{Method} & \textbf{mIoU (\%)} \\
		\midrule
		J-CRF \cite{zheng2014dense} & 14.4 \\
		JH-CRF \cite{zheng2014dense} & 15.1 \\
		PSPNet \cite{zhao2017pyramid} & 36.7  \\
		DeepLab V3 \cite{chen2017rethinking} &  38.1 \\
		\midrule
		Ours (single) & {44.2} \\
		Cerberus & $\textbf{45.3}$ \\
		\bottomrule
	\end{tabular}
	\caption{Attribute quantitative results on NYUd2.}
	\label{tab:attribute}
\end{table}

\begin{table}
	\centering
	\begin{tabular}{@{}lc@{}}
		\toprule
		\textbf{Method} & \textbf{mIoU (\%)} \\
		\midrule
		Roy et al. \cite{roy2016multi} & 49.6 \\
		Roy et al.  (w/ GT) \cite{roy2016multi} & 53.2 \\
		PSPNet \cite{zhao2017pyramid} & 60.4  \\
		DeepLab V3 \cite{chen2017rethinking} & 61.4  \\
		\midrule
		Ours (single) & {65.2} \\
		Cerberus & $\textbf{66.3}$ \\
		\bottomrule
	\end{tabular}
	\caption{Affordance quantitative results on NYUd2.}
	\label{tab:affordance}
\end{table}

\section{Experiment}

\subsection{Comparisons with State-of-the-art Methods}
\textbf{Evaluation details.} We benchmark our multi-task dense prediction model on NYUd2\cite{silberman2012indoor} dataset. NYUd2 contains 1449 RGB-D images of indoor scenes with 40 object categories. \cite{roy2016multi} augments NYUd2 with additional five affordance maps: \emph{sittable}, \emph{walkable}, \emph{lyable}, \emph{reachable} and \emph{movable}. Furthermore, \cite{zheng2014dense} annotates the dataset with 11 additional attribute labels. We train and evaluate Cerberus only with RGB input, using 795 images for training and 654 images for testing. For comparison, we additionally train two widely-used CNN-based dense prediction network PSPNet\cite{zhao2017pyramid} and DeepLab V3 \cite{chen2017rethinking}. Following previous works, we choose the mean intersection over union (mIoU) score as the evaluation metric for both one-label semantic parsing and multi-label affordance/attribute parsing.  

\textbf{Attribute.}
We show our attribute prediction results in Tab.\ref{tab:attribute}. We compare Cerberus against best published results. In contrast to DeepLab v3, our attribute parsing mIoU is significantly  promoted  from 38.1\% to  45.3\%. Besides, we train a single-task attribute parsing transformer denoted as Ours (single), to investigate the effect of joint learning. As shown in Tab.\ref{tab:attribute}, the mIoU of Cerberus outperforms Ours (single) by 1.1\%. This shows that cues from the other two tasks help regularize the shared attention and improve the performance of attribute parsing. 

\begin{figure*}[htbp]
  \centering
  \includegraphics[width=0.8\linewidth]{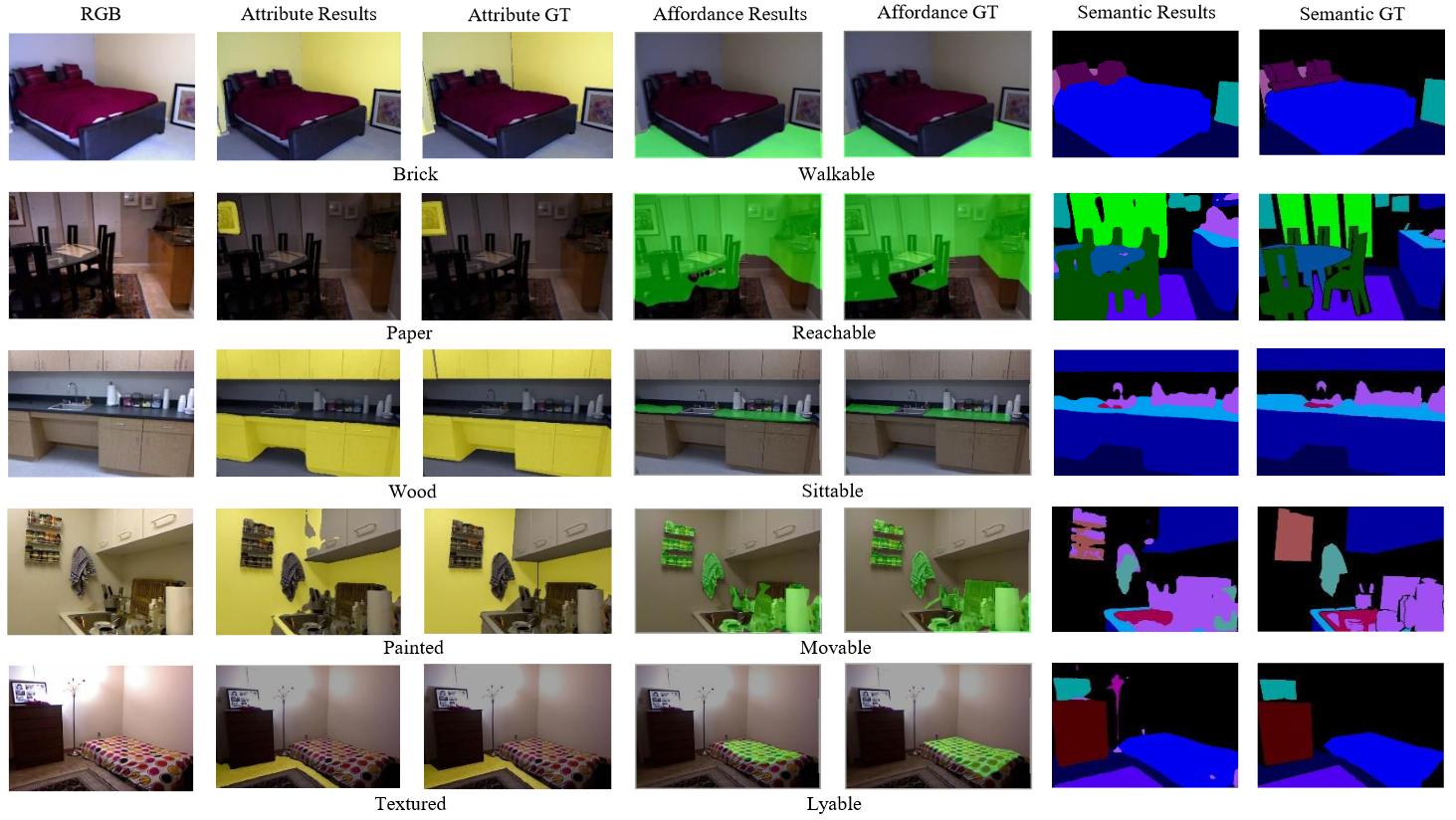}
  \caption{Qualitative prediction results on NYUd2 for three tasks addressed.}
  \label{fig:qualitative} 
\end{figure*}

\begin{figure*}[htbp]
	\centering
	\includegraphics[width=0.8\linewidth]{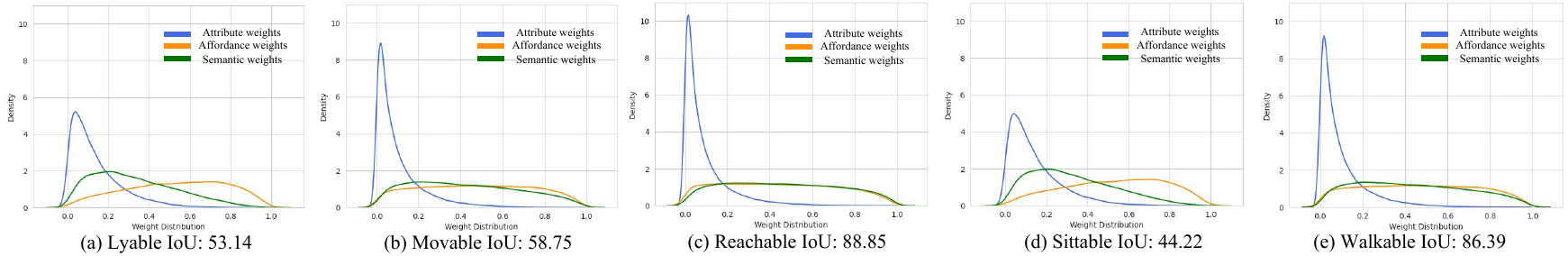}
	\caption{\textbf{Weight distribution} (kernel density estimation) for different affordance classes during the training of Cerberus.}
	\label{fig:Weights} 
\end{figure*}

\textbf{Affordance.}
We provide the performance of affordance prediction in Tab.\ref{tab:affordance}.  These results suggest a large improvement over the previous state-of-the-art. We achieve a 13.1\% boost over Roy et al. (w/ GT) \cite{roy2016multi}, which uses ground truth cues. Similar to attribute, Cerberus also obtains superior performance over separately trained model, verifying the effectiveness of multi-task learning.

\begin{table}
	\centering
	\begin{tabular}{ccc}
		\toprule
		\textbf{Method} & \textbf{Input} & \textbf{mIoU (\%)} \\
		\midrule
		3DGNN \cite{qi20173d} & \multicolumn{1}{|c|} {RGB-D} & \multicolumn{1}{|c} {43.1}\\
		RDF-101 \cite{park2017rdfnet} &\multicolumn{1}{|c|} {RGB-D} & \multicolumn{1}{|c} { 49.1} \\
		ACNet \cite{hu2019acnet} &\multicolumn{1}{|c|} {RGB-D} & \multicolumn{1}{|c} {48.3} \\
		\midrule
		PSPNet \cite{zhao2017pyramid} &  \multicolumn{1}{|c|} {RGB}  & \multicolumn{1}{|c} {43.1} \\
		DeepLab V3 \cite{chen2017rethinking} & \multicolumn{1}{|c|} {RGB} & \multicolumn{1}{|c} {44.7}\\
		OCNet \cite{yuan2018ocnet} &\multicolumn{1}{|c|} {RGB} & \multicolumn{1}{|c} {44.5} \\
		FastFCN \cite{wu2019fastfcn} & \multicolumn{1}{|c|} {RGB} &\multicolumn{1}{|c} {45.4} \\
		VarReg \cite{shi2019scene} & \multicolumn{1}{|c|} {RGB} &\multicolumn{1}{|c} {\textbf{50.7}} \\
		\midrule
		Ours (single) & \multicolumn{1}{|c|} {RGB} &\multicolumn{1}{|c} {48.8} \\
		Cerberus &\multicolumn{1}{|c|} {RGB} & \multicolumn{1}{|c}{50.4}  \\
		\bottomrule
	\end{tabular}
	\caption{Semantic quantitative results on NYUd2.}
	\label{tab:seg}
\end{table}

\begin{table}
	\centering
	\small 
	\begin{tabular}{c|c|ccc}
		\toprule
\textbf{Architecture} & \textbf{Weights} & \textbf{Attr.} & \ \textbf{Aff.} & \textbf{Sem.} \\
		\midrule
	    PSPNet  & single & 36.7 & 60.4 &  43.1\\
	    PSPNet & uniform & 38.3 & 60.3 & 42.4 \\
		PSPNet & optimal &\textbf{38.6} & {$\textbf{61.3}$}  & $\textbf{43.2}$ \\
		\midrule
		DeepLab V3  &single&  38.1 & 61.4 & 44.7 \\
		DeepLab V3 & uniform &  41.1 & 62.5 & 43.6\\
		DeepLab V3 & optimal &$\textbf{42.2}$ &$\textbf{63.2}$ & $\textbf{44.8}$  \\	
		\midrule
		Cerberus &single & {44.2} & 65.2 & 48.8\\
		Cerberus & uniform &  {44.5} & 63.9 & 48.3\\
		Cerberus & optimal & {$\textbf{45.3}$}  & {$\textbf{66.3}$} & {$\textbf{50.4}$} \\
		\bottomrule
	\end{tabular}
	\caption{Quantitative results on NYUd2 with or without using the proposed optimal weights balancing scheme.}
	\label{tab:moo}
\end{table}
\textbf{Semantics}. Tab.\ref{tab:seg} provides the performance of Cerberus on the NYUd2 semantic parsing task. Our model outperforms most of the previous state-of-the-arts and achieves comparable performance with SceneParsing \cite{shi2019scene}. The baseline results reported here are mainly evaluated by \cite{chen2020bi}.  Like the other two tasks, Cerberus outperforms the separately trained model by 1.6\%, indicating that multi-task training is of crucial importance.

\textbf{Qualitative results}. Fig.\ref{fig:qualitative} shows our joint parsing results on NYUd2. Our model can predict attribute, affordance and semantic maps precisely in diverse indoor scenes. With all the predicted semantics and properties, one may have an internal image of the scene even without seeing the original RGB image. From the prediction results in the third row of Fig.\ref{fig:qualitative}, we know there is a cabinet made of wood, and its surface is \emph{sittable}. And in the fourth row, we could see there are many \emph{movable} objects hanging on the painted wall. These results are beneficial to a range of applications, like intelligent service robots and augmented reality.




\subsection{Experiments on Optimal Weights} 

\textbf{Optimal weights effectively balance Cerberus.} We conduct experiments to understand the effect of using optimal weights. For comparison, we train a Cerberus using a uniformly weighted multi-task loss, and the results are shown in Tab.\ref{tab:moo}. We also conduct experiments with two CNN-based models under the same settings to further show the role of optimal weights. It is clear that training with optimal weights gets superior performance. Specifically, for Cerberus, the mIoU of attribute is increased by 0.8\%, affordance performance increases 2.3\% and mIoU of semantic is also raised by 2.1\%. Notably, using a uniformly weighted loss even results in lower performance on certain sub-tasks, when compared with separately trained models. This indicates that there are inconsistent gradient directions between tasks during training and using uniform weights cannot successfully leverage task affinity to resolve them, while optimal weights can better unleash the power of multi-task training to achieve that.

\textbf{Optimal weights bias towards tougher tasks}. To investigate how optimal weights benefit Cerberus training, we visualize the weight distributions during training in Fig.\ref{fig:Weights}, taking affordance as an example. For every affordance label, we collect a subset of the train set which contains all the samples in which the label exists. We save the corresponding optimal weights when encountering these samples. As shown in Fig.\ref{fig:Weights}, different labels correspond to different distributions. For \emph{Lyable} and \emph{Sittable}, the affordance weights are higher than in other affordance distributions. And these two classes have lower mIoU results. This reveals that when training tougher sub-tasks, optimal weights tend to be higher. And through dynamic balancing between different sub-tasks, Cerberus achieves superior performance. 

\begin{figure*}[htbp]
  \centering
  \includegraphics[width=0.7\linewidth]{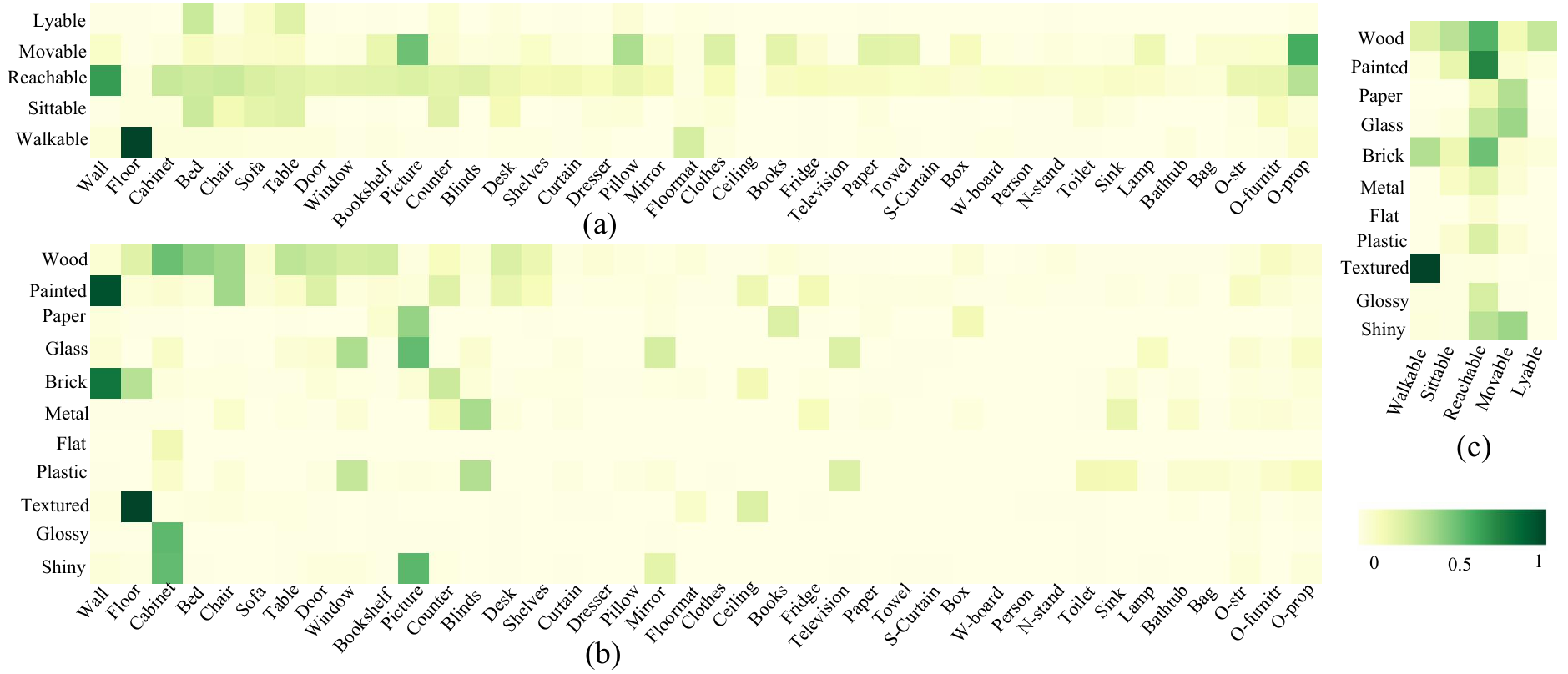}
  \caption{\textbf{Visualization of task affinity}. We calculate mIoU between different sub-task concepts, and visualize them with color maps.}
  \label{fig:task} 
\end{figure*}

\begin{table}
	\centering
	\begin{tabular}{ccccc}
		\toprule
		\textbf{Model}  & \textbf{Attribute} & \ \textbf{Affordance} & \textbf{Semantic} \\
		\midrule
		single (at) & \multicolumn{1}{|c} {38.3} & n/a & n/a \\
		Uniform (at) & \multicolumn{1}{|c} {43.2} & 65.2 & 46.8 \\
		Cerberus (at) & \multicolumn{1}{|c} {$\textbf{44.1}$} & 65.2 & 47.1 \\
		\midrule
		single (af) & \multicolumn{1}{|c} {n/a} & 60.9 & n/a \\
		Uniform (af) & \multicolumn{1}{|c} {44.7} & $\textbf{64.2}$ & 46.9 \\
		Cerberus (af) & \multicolumn{1}{|c} {44.6} & 64.1 & 47.2 \\
		\midrule
		single (sem) & \multicolumn{1}{|c} {n/a} & n/a & 23.9 \\
		Uniform (sem) & \multicolumn{1}{|c} {44.3} & 63.1 & 37.6 \\
		Cerberus (sem) & \multicolumn{1}{|c} {44.7} & 65.9 & $\textbf{42.7}$ \\
		\bottomrule
	\end{tabular}
	\caption{1\% weakly-supervised experiments on NYUd2.}
	\label{tab:weak1}
\end{table}

\begin{table}
	\centering
	\begin{tabular}{cccc}
		\toprule
		\textbf{Method} & \textbf{Attribute} & \ \textbf{Affordance} & \textbf{Semantic} \\
		\midrule
		single (at) & \multicolumn{1}{|c} {37.1} & n/a & n/a \\
		Uniform (at) & \multicolumn{1}{|c} {44.1} & 63.3 & 47.7 \\
		Cerberus (at)& \multicolumn{1}{|c} {$\textbf{44.2}$} &65.4 & 46.3\\
		\midrule
		single (af) & \multicolumn{1}{|c} {n/a} & 58.8 & n/a \\
		Uniform (af) & \multicolumn{1}{|c} {43.2} & 62.5 & 48.9 \\
		Cerberus (af)& \multicolumn{1}{|c} {44.5} & {$\textbf{63.5}$} & 48.4 \\
		\midrule
		single (sem) & \multicolumn{1}{|c} {n/a} & n/a & 20.2 \\
		Uniform (sem) & \multicolumn{1}{|c} {44.5} & 65.3 & 36.5 \\
		Cerberus (sem)& \multicolumn{1}{|c} {43.8} & 65.2  & {$\textbf{39.9}$} \\
		\bottomrule
	\end{tabular}
	\caption{0.5\% weakly-supervised experiments on NYUd2.}
	\label{tab:weak2}
\end{table}
 
\begin{table}
	\centering
	\begin{tabular}{cccc}
		\toprule
		\textbf{Method} & \textbf{Attribute} & \ \textbf{Affordance} & \textbf{Semantic} \\
		\midrule
		single (at) & \multicolumn{1}{|c} {36.4} & n/a & n/a \\
		Uniform (at)& \multicolumn{1}{|c} {41.8} & 64.3  & 47.8 \\
		Cerberus (at)& \multicolumn{1}{|c} {$\textbf{43.5}$} &65.3 &49.3 \\
		\midrule
		single (af) & \multicolumn{1}{|c} {n/a} &57.5  & n/a \\
		Uniform (af)& \multicolumn{1}{|c} {45.0} & 62.4  & 46.2 \\
		Cerberus (af)& \multicolumn{1}{|c} {43.9} & {$\textbf{64.1}$} & 48.7 \\
		\midrule
		single (sem) & \multicolumn{1}{|c} {n/a} & n/a & 18.6 \\
		Uniform (sem)& \multicolumn{1}{|c} {43.3} & 64.4  & 32.3 \\
		Cerberus (sem)& \multicolumn{1}{|c} {44.3} & 65.4  & {$\textbf{39.1}$} \\
		\bottomrule
	\end{tabular}
	\caption{0.1\% weakly-supervised experiments on NYUd2.}
	\label{tab:weak3}
\end{table}

\subsection{Towards a Deeper Understanding of Cerberus}

\textbf{Task Affinity.} Inspired by the semantic transfer technique in \cite{zhao2017physics}, we conduct experiments to explore the sub-task relationships. Due to the fact of weakly aligned data, we can't use the annotation masks directly. Instead we use aligned predictions generated from a single image to explore task affinity learnt by Cerberus. And we quantify the affinity between label $k_1$ and $k_2$ by calculating the intersection over union score (IoU):
\begin{align}
{\rm IoU = \frac{M_{k_1} \cap M_{k_2}}{M_{k_1} \cup M_{k_2}} }
\end{align}
$\rm M_{k_i}$ is the predicted mask for label $k_i$. We calculate the mIoU of all label pairs between different tasks on NYUd2 test set, and the task affinities are visualized in Fig.\ref{fig:task}.

\begin{figure*}[htbp]
  \centering
  \includegraphics[width=0.74\linewidth]{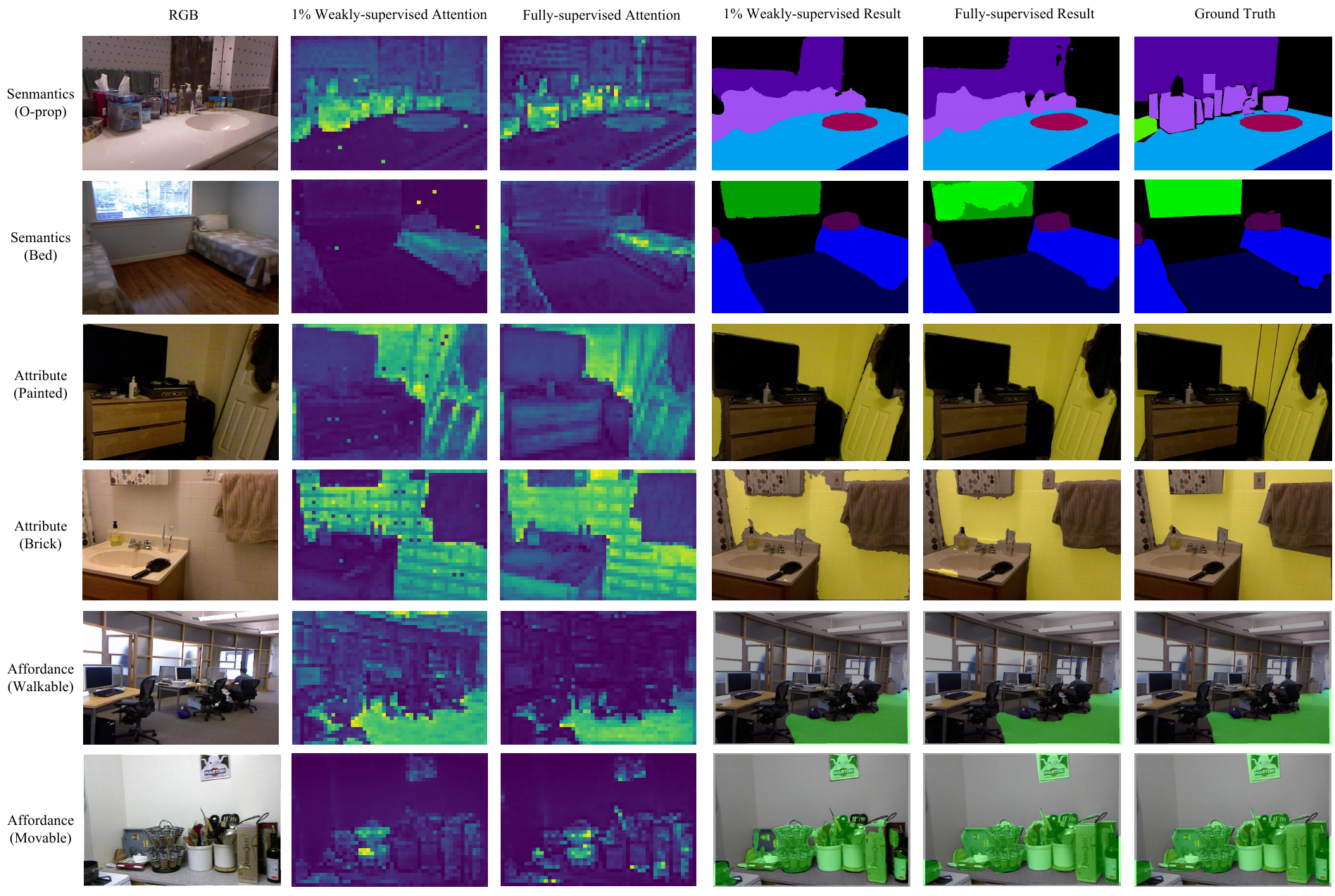}
  \caption{\textbf{Visualization of attention weights}. We analyze the self-attention weights of the readout token on different task heads. Interestingly, in the weakly-supervised setting, learned attention weights still well align with ground truth regions.}
  \label{fig:attention} 
\end{figure*}

As shown in the figure, the results of different pairs vary a lot but are still reasonable. For example, \emph{Textured}, \emph{Floor} and \emph{Walkable} have high mIoU with each other, since floors are usually textured and walkable. And walls are often painted, leading to a high correlation between them. The high mIoU pairs in semantic-attribute affinity are also in line with common sense: Windows and pictures are usually made of glass, while cabinets and chairs are typically made of wood. The task affinities reveal that the three tasks are not independent of each other, and learning one of them may help the other two. That's the potential reason why our Cerberus out-performs separately trained models. 

\textbf{Cerberus performs well under weak supervision}. Inspired by task affinity, we conduct a set of experiments to further unleash the potential of Cerberus. We train a set of Cerberus with one task supervised by $0.1\%-1\%$ annotation while the other two by full supervision. For comparison, we also train single-task models with $0.1\%-1\%$ annotation. We use a random mask to select the annotation. The results are shown in Tab. \ref{tab:weak1}, \ref{tab:weak2} and \ref{tab:weak3}. Our weakly-supervised Cerberus models outperform single-task models, and multi-task models trained with uniformly-weighted losses. For attribute and affordance, the weakly-supervised Cerberus are only slightly worse than the fully supervised Cerberus. Specially,  on 0.1\% weakly-supervised semantic task, Cerberus outperforms separately trained model by 20.5\%. 

\textbf{Shared attention facilitates weakly-supervised learning.} In order to explore how Cerberus actually helps with multi-task learning and weakly-supervised learning, we visualize the attention maps of readout token in Fig.\ref{fig:attention}. Though the readout token is not grounded in the input image, it can aggregates information from other tokens. As can be seen, different heads of the last attention layer aggregate different task-specific features. And we observe that, even for weakly-supervised settings, Cerberus still successfully generates attention maps well aligned with ground truth regions, which appears very similar to the fully supervised attention maps. We believe that this is because that related sub-tasks help attention learning with little annotation. We consider this as a human-like learning capability: one can achieve weakly-supervised learning with the help of other related tasks. For example, if we know what is window, we can easily learn how glass looks like. That's why our attention-based model achieves strong performance under weak supervision.

\section{Conclusion}
In this paper, we propose a novel multi-task dense prediction transformer named Cerberus to parse semantics, affordance and attribute jointly. We successfully train our model from weakly-aligned data using a weight balancing framework to unleash the power of multi-task learning. Cerberus achieves state-of-the-art performance on all tasks and strong results under weak supervision (using as low as 0.1\% annotation). We observe task affinity consistent with common sense and further demonstrate that shared attention between tasks facilitates weakly-supervised learning.

{\small
\bibliographystyle{ieee_fullname}
\bibliography{egbib}
}

\newpage

\section{Appendix}

In this supplementary material, we provide more experiment results to show the effectiveness of Cerberus and validate our assumption that task affinity helps Cerberus learn under weak supervision. In the following sections, we first provide per-category evaluation results for semantic, affordance and attribute parsing respectively. Then we provide more visualization results for both parsing and attention. Shared attention weights reveal the potential reason behind strong weakly-supervised learning performance.

\subsection{Per-Category Evaluation}
Tab.~\ref{tab:sem}, \ref{tab:aff} and \ref{tab:att} demonstrate our per-category semantic, affordance and attribute mIoU results on NYUd2\cite{silberman2012indoor} respectively. \emph{Single} denotes the separately trained model for different tasks. And the percentage represents the amount of annotation used by weak supervision. It is manifested that our Cerberus achieves the best performance in most categories. And it also achieves superior performance than \emph{Single} under the same amount of weak supervision.

Notably, the mIoUs of certain semantic labels are 0 when using a weakly supervised \emph{Single} model. And we observe that these categories often appear as small regions, like paper, or have a complicated internal structure, like person. For these categories, the number of randomly sampled pixels is too small to provide enough information for semantic parsing. Hence these sub-tasks cannot be learned under weak supervision effectively. However, when using a Cerberus model, the mIoUs of these sub-tasks are greater than zero, which verifies task affinity does help weakly-supervised learning, especially for those hard sub-tasks.

\subsection{More Qualitative Results}

We show more qualitative results in Fig.~\ref{fig:more}.  We choose various indoor scenes with different semantic, affordance, and attribute labels. As shown in Fig.~\ref{fig:more}, Cerberus achieves precise attribute, affordance and semantic parsing in all these scenes. For example, in row 6 of Fig.~\ref{fig:more}, though both sides of the room are white, Cerberus can precisely distinguish the difference between blinds and walls. Accurate parsing is also observed in the \emph{Painted} attribute results.  And in the affordance results of row 6, Cerberus can even identify regions on the bed that are too far away to sit on.  Considering the diversity of scenes, we believe Cerberus is accurate enough for various applications including augmented reality and intelligent service robots . 

\subsection{More Attention Visualization}

Fig.~\ref{fig:attn} provides more attention maps and corresponding parsing results. We train three Cerberus with one task supervised by $1\%$ annotation while the other two by full supervision, and compare them with fully-supervised Cerberus.  As shown in the figure, attention maps focus on those regions that correspond to parsing results. For example, in row 8 of Fig.~\ref{fig:attn}, the attention weights exactly focus on all movable objects, including objects on shelves, objects on the cabinets, and even the painting on the wall. Meanwhile, for the weakly-supervised model, we still can find attention maps showing the corresponding features, which appear very similar to the fully supervised attention maps. We believe that this is because those related sub-tasks help attention learning with little annotation. And with shared attention, we achieve strong results under weak supervision.

\section{Computation Cost and Failure case}
In Fig. \ref{fig:pie}, the computation cost is visualized as the training time distribution pie chart, which shows solving optimal weights is efficient and only takes 10\% of the training time. 

As shown in Fig. \ref{fig:fc} for the failure case, due to the high affinity between \emph{walkable} and \emph{textured}, Cerberus is affected by the shared representation and performs worse than a separately trained attribute network.

\begin{table*}
\small
  \centering
  \begin{tabular}{c|cccccccccc|c}
    \toprule
    \textbf{Method} & \rotatebox{90}{Wall} & \rotatebox{90}{Floor} & \rotatebox{90}{Cabinet} & \rotatebox{90}{Bed}
 & \rotatebox{90}{Chair} & \rotatebox{90}{Sofa} & \rotatebox{90}{Table} & \rotatebox{90}{Door} & \rotatebox{90}{Window} & \rotatebox{90}{Bookshelf} & mIoU \\
    \midrule
    Single (1\%) & 71.5 & 82.0 & 53.9 & 67.4 & 59.0 & 60.8 & 46.2 & 20.8 & 43.3 & 42.4 & 23.9\\
    Single (0.5\%)  &71.6 & 81.2 & 53.3 & 63.2 & 57.4 & 55.1 & 44.3 & 15.7 & 30.5 & 39.7 & 20.2\\
    Single (0.1\%) &68.3 & 80.3 & 45.1 & 64.0 & 52.2 & 58.5 & 42.8 & 5.7 & 28.6 & 38.2 & 18.6  \\
    \midrule
    Cerberus (1\%) & 77.7 & 87.3 & 63.9 & 71.0 & 64.2 & 65.4 & 51.7 & 37.5 & 50.1 & 44.3 & 42.7\\
    Cerberus (0.5\%)  & 75.9 & 87.0 & 63.6 & 70.7 & 65.5 & 67.5 & 50.5 & 33.1 & 51.0 & 41.9 & 39.9\\
    Cerberus (0.1\%) &79.4 & 86.8 & 62.3 & 70.1 & 64.2 & 66.2 & 46.8 & 41.7 & 48.8 & 43.0 & 39.1\\
    \midrule
    Single & 80.0 & 88.0 & 61.8 & 72.7 & \textbf{66.6} & \textbf{67.8} & 53.2 & \textbf{43.0} & \textbf{52.5} & 43.8 & 48.8\\
    Uniform & \textbf{80.8} & 87.7 & 62.3 & 71.7 & 65.2 & 64.1 & 49.8 & 42.3 & 50.4 & \textbf{46.0} & 48.3\\
    Cerberus & 79.6 & \textbf{88.7} & \textbf{64.7} & \textbf{72.8} & 64.7 &65.5 & \textbf{54.4} & 41.1 & 48.0 & 44.5  & \textbf{50.4}\\
    \toprule
     \textbf{Method} & \rotatebox{90}{Picture} & \rotatebox{90}{Counter} & \rotatebox{90}{Blinds} & \rotatebox{90}{Desk}
 & \rotatebox{90}{Shelves} & \rotatebox{90}{Curtain} & \rotatebox{90}{Dresser} & \rotatebox{90}{Pillow} & \rotatebox{90}{Mirror} & \rotatebox{90}{Floormat}  \\
     \midrule
    Single (1\%) & 51.5 & 56.0 & 55.3 & 13.4 & 1.4 & 51.5 & 34.2 & 26.8 & 0.3 & 15.6\\
    Single (0.5\%) &47.0 & 56.7 & 52.7 & 3.6 & 3.0 & 17.6 & 6.9 & 19.5 & 1.1 & 0.4 \\
    Single (0.1\%) &47.3 & 55.6 & 48.1 & 0.0 & 0.0 & 10.8 & 1.1 & 10.1 & 0.0 & 8.3\\
    \midrule
    Cerberus (1\%) & 58.4 & 67.0 & 64.8 & 28.2 & 14.7 & 59.0 & 47.4 & 43.5 & 42.4 & 42.3\\
    Cerberus (0.5\%) &56.1 & 65.7 & 63.1 & 28.0 & 10.8 & 63.3 & 47.1 & 45.4 & 46.1 & 35.3 \\
    Cerberus (0.1\%) & 58.1 & 64.6 & 64.2 & 26.7 & 12.8 & 60.5 & 41.1 & 43.5 & 43.3 & 40.3\\
    \midrule
    Single & 58.1 & 66.7 & 61.8 & 27.0 & \textbf{17.7} & 58.4 & 47.6 & 43.1 & \textbf{49.3} & 39.8 \\
    Uniform &  57.1 & 68.8 & \textbf{65.3} & 26.6 & 16.9 & 60.0 & 51.9 & 35.2 & 49.0 & 45.9\\
    Cerberus & \textbf{60.5} & \textbf{70.6} & 63.8 & \textbf{28.3} & \textbf{17.7} & \textbf{64.7} & \textbf{54.8} & \textbf{44.2} & 47.4 & \textbf{46.2}  \\
    \toprule
    \textbf{Method} & \rotatebox{90}{Clothes} & \rotatebox{90}{Ceiling} & \rotatebox{90}{Books} & \rotatebox{90}{Fridge}
 & \rotatebox{90}{Television} & \rotatebox{90}{Paper} & \rotatebox{90}{Towel} & \rotatebox{90}{S-curtain} & \rotatebox{90}{Box} & \rotatebox{90}{W-board}  \\
     \midrule
    Single (1\%) & 8.7 & 0.1 & 1.0 & 0.0 & 43.4 & 0.0 & 0.0 & 0.0 & 0.0 & 0.0 \\
    Single (0.5\%) &5.5 & 0.6 & 0.9 & 0.0 & 39.6 & 0.0 & 0.0 & 0.0 & 0.0 & 0.0\\
    Single (0.1\%) &5.6 & 0.0 & 0.0 & 0.0 & 35.4 & 0.0 & 0.0 & 0.0 & 0.0 & 0.0 \\
    \midrule
    Cerberus (1\%) & 21.5 & 56.2 & 18.7 & 21.7 & 65.7 & 5.6 & 36.1 & 16.1 & 9.4 & 38.7\\
    Cerberus (0.5\%) &21.7 & 57.4 & 8.6 & 0.0 & 61.0 & 0.2 & 36.6 & 13.5 & 3.0 & 65.6  \\
    Cerberus (0.1\%) & 16.8 & 55.8 & 21.3 & 44.2 & 62.5 & 1.5 & 5.0 & 17.0 & 3.0 & 0.2  \\
    \midrule
    Single & \textbf{23.0} & 56.5 & \textbf{36.8} & \textbf{66.4} & 60.3 & 30.7 & 48.2 & 33.7 & 15.9 & 74.8 \\
    Uniform & 22.9 & 59.9 & 35.2 & 58.9 & 57.8 & 32.2 & 44.8 & 38.6 & 14.5 & \textbf{80.5}  \\
    Cerberus & 22.4 & \textbf{60.2} & 34.8 & 64.9 & \textbf{65.5} & \textbf{38.2} & \textbf{50.3} & \textbf{45.2} & \textbf{18.7} & 74.2  \\
    \toprule
    \textbf{Method} & \rotatebox{90}{Person} & \rotatebox{90}{N-stand} & \rotatebox{90}{Toilet}
 & \rotatebox{90}{Sink}
 & \rotatebox{90}{Lamp} & \rotatebox{90}{Bathtub} & \rotatebox{90}{Bag} & \rotatebox{90}{O-str} & \rotatebox{90}{O-furnitr} & \rotatebox{90}{O-prop}  \\
     \midrule
    Single (1\%) & 0.0 & 0.0 & 0.0 & 0.0 & 0.0 & 0.0 & 0.0 & 9.7 & 6.9 & 33.1\\
    Single (0.5\%)  &0.0 & 0.0 & 0.0 & 0.0 & 0.0 & 0.0 & 0.0 &3.3 & 2.9 & 24.5\\
    Single (0.1\%) &0.0 & 0.0 & 0.0 & 0.0 & 0.0 & 0.0 & 0.0 & 6.7 & 1.1 & 31.7\\
    \midrule
    Cerberus (1\%) & 74.6 & 9.1 & 61.2 & 47.3 & 34.9 & 24.7 & 0.0 & 27.3 & 20.6 & 38.9\\
    Cerberus (0.5\%) &76.0 & 0.3 & 63.1 & 35.3 & 0.1 & 2.3 & 0.0 & 26.9 & 20.5 & 37.6   \\
    Cerberus (0.1\%) & 74.4 & 0.6 & 56.6 & 42.9 & 0.0 & 9.6 & 0.0 & 27.5 & 21.4 & 37.2\\
    \midrule
    Single & \textbf{82.1} & 42.8 & 64.1 & 46.8 & 42.0 & 34.9 & 0.0 & \textbf{33.4} & \textbf{22.1} & 40.2 \\
    Uniform &  70.1 & 43.1 & 62.6 & 49.7 & 40.1 & 32.9 & 0.0 & 31.6 & 20.2 & 40.0 \\
    Cerberus & 74.7 & \textbf{43.3} & \textbf{66.4} & \textbf{53.7} & \textbf{44.3} & \textbf{37.2} & \textbf{6.0} & 33.1 & 21.8 & \textbf{40.6}  \\
    \bottomrule
  \end{tabular}
  \caption{Per-category semantic parsing results on NYUd2.}
  \label{tab:sem}
\end{table*}

\begin{table*}
  \centering
  \small
  \begin{tabular}{c|ccccc|c}
    \toprule
    \textbf{Method} & Lyable & Movable & Reachable & Sittable & Walkble & mIoU \\
    \midrule
    Single (1\%)  & 40.1 & 55.1 & 87.4 & 41.1 & 81.1 & 60.9\\
    Single (0.5\%)  & 37.1 & 53.4 & 86.2 & 36.6 & 80.6 & 58.8 \\
    Single (0.1\%) &39.8 & 45.5 & 84.9 & 39.6 & 77.1 & 57.5\\
    \midrule
    Cerberus (1\%) &51.3 & 57.3 & 87.9 & 41.1 & 82.9& 64.1\\
    Cerberus (0.5\%)  &49.1 & 57.0 & 87.9 & 39.5 & 84.0& 63.5\\
    Cerberus (0.1\%) &51.3 & 57.3 & 87.8 & 41.1 & 82.9& 64.1\\
    \midrule
    Single &  51.4 & 57.5 & 87.7 & 43.4 & 85.9 &65.2\\
    Uniform &47.2 & 55.8 & 88.1 & 43.5 & 85.1&63.9\\
    Cerberus & \textbf{53.1} & \textbf{58.7} & \textbf{88.9} & \textbf{44.2} & \textbf{88.3}&\textbf{66.3}  \\
    \bottomrule
  \end{tabular}
  \caption{Per-category affordance parsing results on NYUd2.}
  \label{tab:aff}
\end{table*}

\begin{table*}
  \centering
  \small
  \begin{tabular}{c|ccccccccccc|c}
    \toprule
    \textbf{Method} & Wood & Painted & Paper & Glass & Brick & Metal & Flat & Plastic & Textured & Glossy & Shiny & mIoU\\
    \midrule
    Single (1\%) &46.2& 64.3& 22.0& 37.8& 45.4& 13.2& 0.0& 27.7& 72.4& 46.4& 46.0 & 38.3 \\
    Single (0.5\%) &49.8& 62.5& 10.6& 36.5& 44.1& 12.0& 0.0& 25.5& 71.9& 48.4& 46.6 &37.1\\
    Single (0.1\%)&47.4& 63.8& 8.5& 37.7& 45.6& 11.6& 0.0& 25.8& 67.0& 46.8& 45.9 &36.4\\
    \midrule
    Cerberus (1\%) &52.0& 67.5& 33.2& \textbf{45.5} & 50.1& 21.1& 3.6& 30.8& 76.2& 51.0& 53.9 &44.1 \\
    Cerberus (0.5\%) &52.9& 66.8& 34.5& 45.4& 50.2& 21.0& 4.2& 30.4& 75.4& 51.2& \textbf{54.1} & 44.2\\
    Cerberus (0.1\%) &52.4& 67.3& 27.0& 45.4& 52.5& 21.9& 3.6& 35.2& 74.8& 49.2& 49.7 & 43.5\\
    \midrule
    Single &52.2&66.8&30.7&44.6&\textbf{52.6}&20.8&2.5&\textbf{35.3}&75.6&51.5&54.0 & 44.2 \\
    Uniform &\textbf{54.5}& 67.8& 29.1& 43.2& 51.7& 25.0& \textbf{6.2}& 31.1& \textbf{76.4}& 50.8& 53.8 & 44.5\\
    Cerberus & 54.3 & \textbf{68.1} & \textbf{36.2} & 45.3 & 51.9 &\textbf{25.1} & 5.4 & 31.9 & 74.5 & \textbf{51.8} & \textbf{54.1} &\textbf{45.3} \\
    \bottomrule
  \end{tabular}
  \caption{Per-category attribute parsing results on NYUd2.}
  \label{tab:att}
\end{table*}

\begin{figure*}[h]
  \centering
    \label{fig:subfig:subfig-a} 
    \includegraphics[width=1\linewidth]{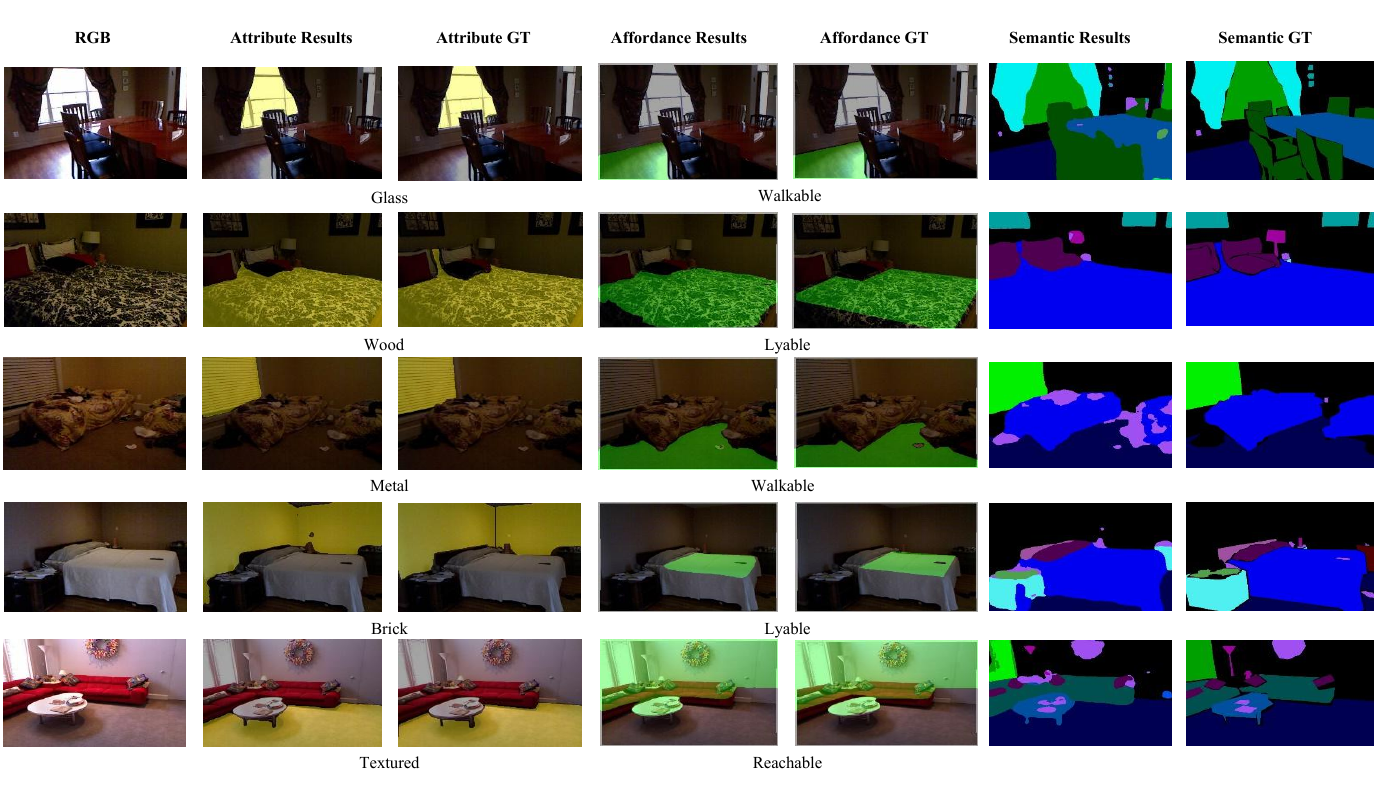}
      \caption{More qualitative results. }
\end{figure*}

\begin{figure*}[t]
  \centering
  \ContinuedFloat
    \label{fig:subfig:subfig-b} 
    \includegraphics[width=1\linewidth]{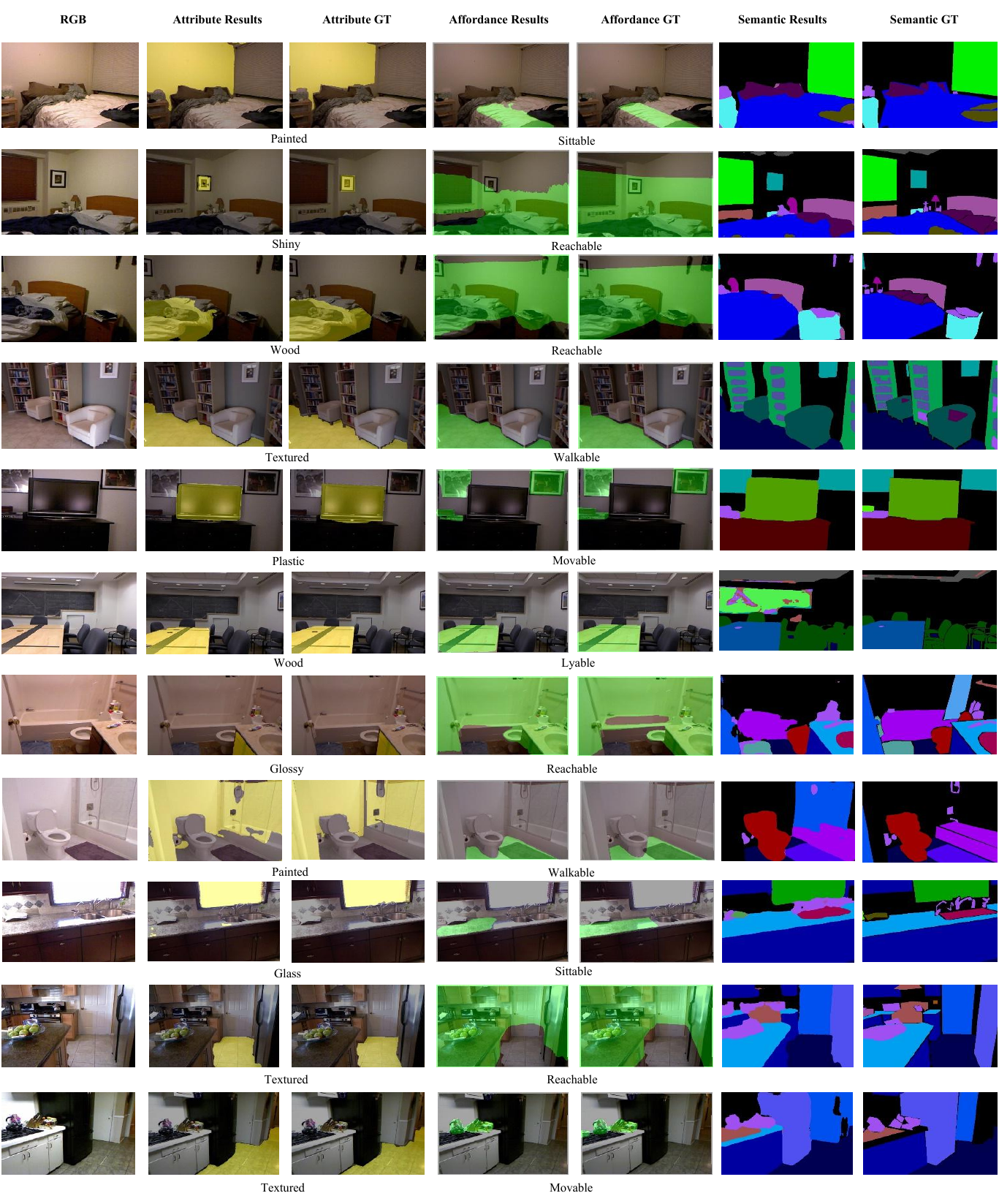}
  \caption{More qualitative results (cont.).}
  \label{fig:more} 
\end{figure*}

\begin{figure*}[t]
  \centering
  \includegraphics[width=1\linewidth]{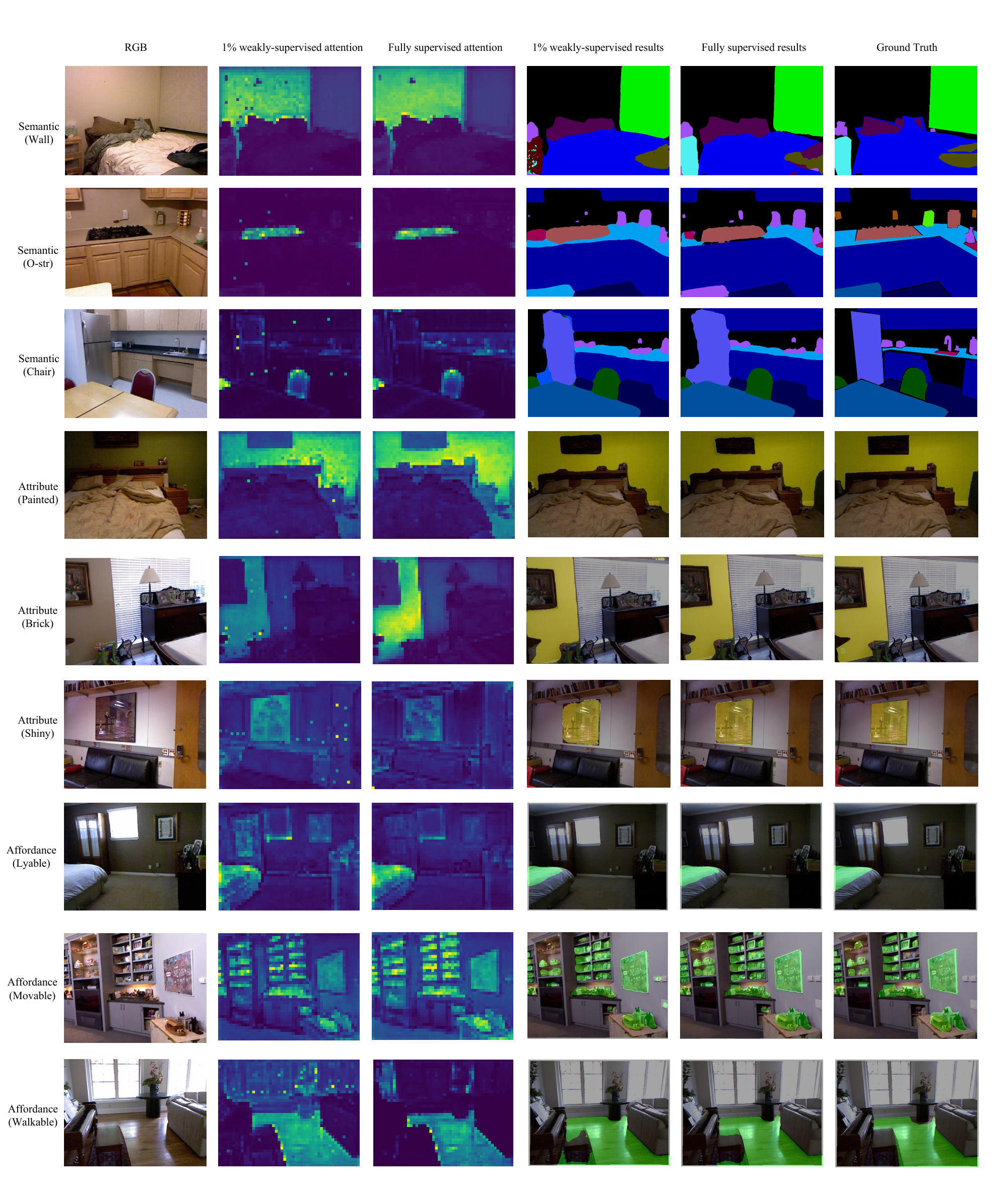}
  \caption{More attention visualizations. }
  \label{fig:attn} 
\end{figure*}

\begin{figure*}[t]
  \centering
  \includegraphics[width=0.7\linewidth]{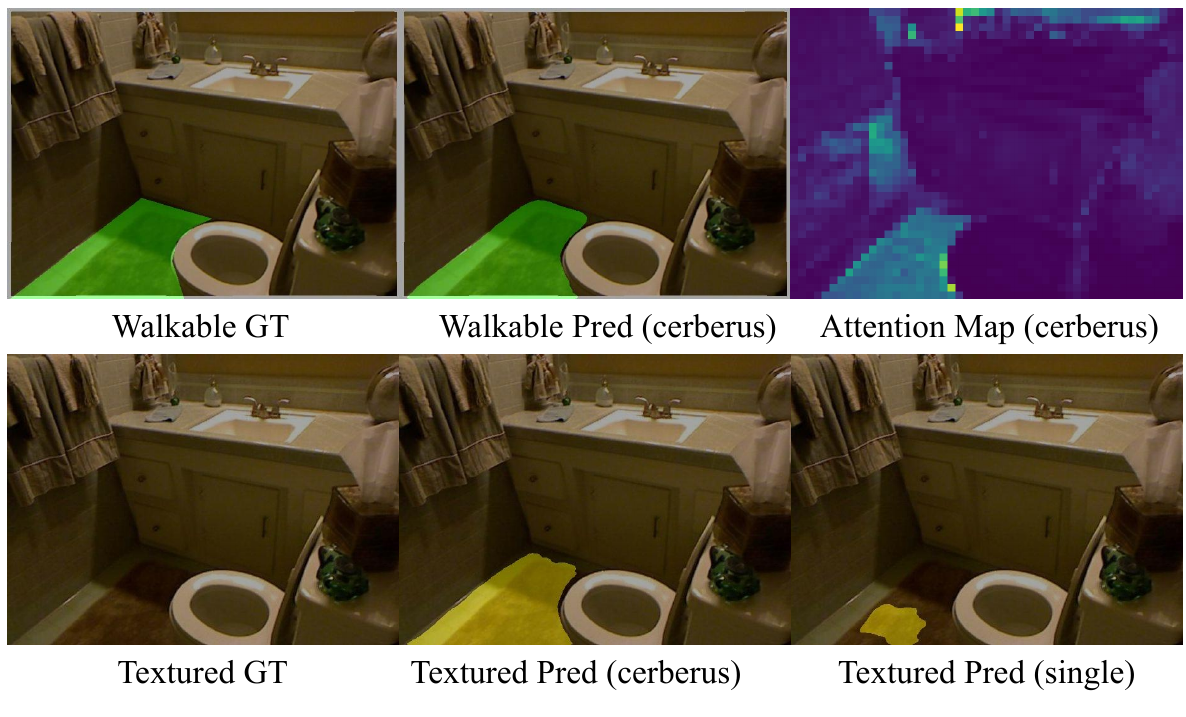}
  \caption{Failure cases. }
  \label{fig:fc} 
\end{figure*}

\begin{figure*}[t]
  \centering
  \includegraphics[width=0.65\linewidth]{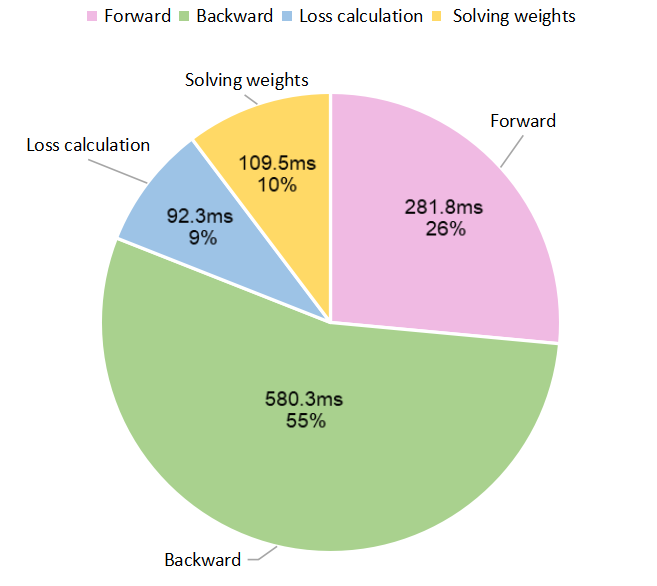}
  \caption{Computation cost distribution pie chart. }
  \label{fig:pie} 
\end{figure*}

\end{document}